\pdfoutput=1

\documentclass[11pt]{article}

\usepackage{acl}

\usepackage{times}
\usepackage{latexsym}

\usepackage[T1]{fontenc}

\usepackage[utf8]{inputenc}

\usepackage{microtype}

\usepackage{inconsolata}

\usepackage{graphicx}

\usepackage{enumitem}
\usepackage{booktabs}
\usepackage{cleveref}

\newcommand\myparagraph[1]{
\vskip 0.05in 
\noindent{\bf {#1}}}

\newcommand{\systemname}{Co-DETECT}
\newcommand{\stepone}{\raisebox{-0.4ex}{\includegraphics[width=0.9em]{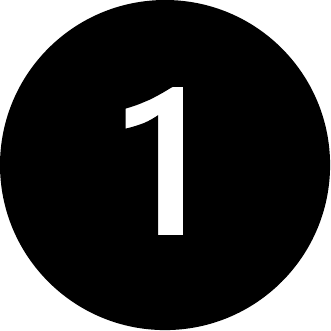}}}
\newcommand{\steptwo}{\raisebox{-0.4ex}{\includegraphics[width=0.9em]{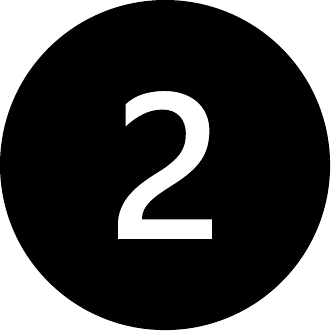}}}

\title{Co-DETECT: Collaborative Discovery of Edge Cases in Text Classification}

\author{
 \textbf{Chenfei Xiong\textsuperscript{$Z$}\thanks{Equal contributions.}}\quad
 \textbf{Jingwei Ni\textsuperscript{$E$ $Z$}\footnotemark[1]}\quad
 \textbf{Yu Fan\textsuperscript{$E$}\footnotemark[1]}\quad 
 \textbf{Vilém Zouhar\textsuperscript{$E$}}\quad \\
 \textbf{Donya Rooein\textsuperscript{$B$ $E$}}\quad 
 \textbf{Lorena Calvo-Bartolomé\textsuperscript{$C$}}\quad
 \textbf{Alexander Hoyle\textsuperscript{$E$}}\quad \\
 \textbf{Zhijing Jin\textsuperscript{$T$}}\quad 
 \textbf{Mrinmaya Sachan\textsuperscript{$E$}}\quad
 \textbf{Markus Leippold\textsuperscript{$Z$}}\quad \\
 \textbf{Dirk Hovy\textsuperscript{$B$}}\quad 
 \textbf{Mennatallah El-Assady\textsuperscript{$E$}}\quad
 \textbf{Elliott Ash\textsuperscript{$E$}}\quad
\\
 \textsuperscript{$E$}ETH Zürich\quad
 \textsuperscript{$Z$}University of Zürich\quad \\
 \textsuperscript{$B$}Bocconi University\quad 
 \textsuperscript{$T$}University of Toronto\quad
 \textsuperscript{$C$}Universidad Carlos III
\\
 \texttt{\{\href{mailto:jingni@ethz.ch}{\color{black} jingni}, \href{mailto:yufan@ethz.ch}{\color{black} yufan}, \href{mailto:ashe@ethz.ch}{\color{black} ashe}\}@ethz.ch}
}

\DeclareUnicodeCharacter{202F}{\,}
\begin{document}
\maketitle

\begin{figure*}[ht]
    \centering
    \includegraphics[width=\textwidth]{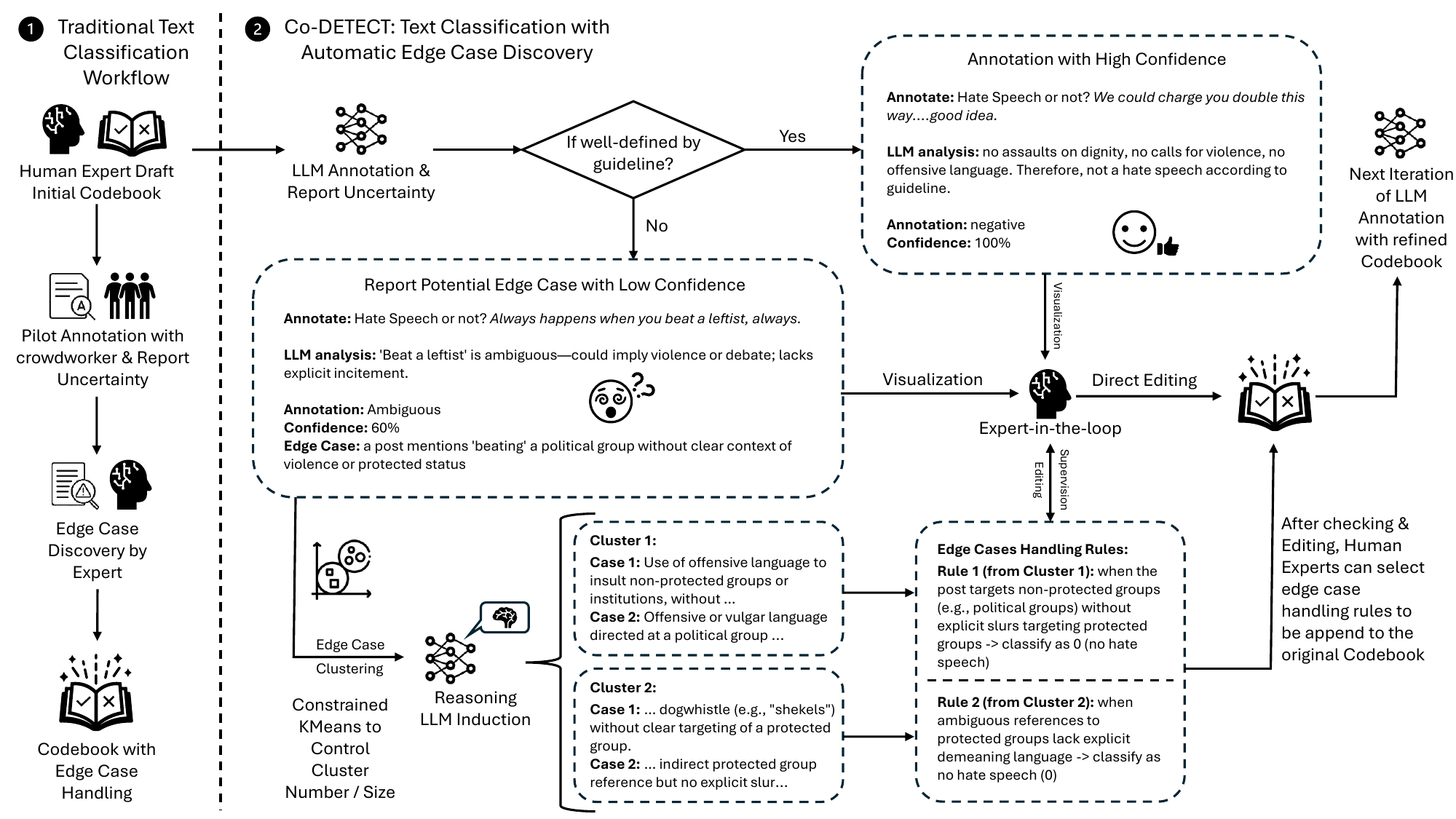}
    \vspace{-2em}
    \caption{\stepone\space traditional workflow of text annotation, where experts rely on their own or crowdworkers to identify edge cases and update codebook based on the discovered prevalent edge cases. \steptwo\space Co-DETECT mixed-initiative workflow of edge case discovery, where LLMs propose prevalent and representative edge cases and the visual interface assists human expert to verify the proposed edge cases.}
    \vspace{-0.7em}
    \label{fig:tf}
\end{figure*}
\begin{abstract}
We introduce \systemname\space (\underline{Co}llaborative \underline{D}iscovery of \underline{E}dge cases in \underline{TE}xt \underline{C}lassifica\underline{T}ion), a novel mixed-initiative annotation framework that integrates human expertise with automatic annotation guided by large language models (LLMs). \systemname\space starts with an initial, sketch-level codebook and dataset provided by a domain expert, then leverages the LLM to annotate the data and identify edge cases that are not well described by the initial codebook. %
Specifically, \systemname\space flags challenging examples, induces high-level, generalizable descriptions of edge cases, and assists user in incorporating edge case handling rules to improve the codebook. This iterative process enables more effective handling of nuanced phenomena through compact, generalizable annotation rules. Extensive user study, qualitative and quantitative analyses prove the effectiveness of \systemname{}.%
\footnote{Frontend: \href{https://codetect.vercel.app/}{codetect.vercel.app}; code and demonstration video: \href{https://github.com/EdisonNi-hku/Co-DETECT/}{github.com/EdisonNi-hku/Co-DETECT}}

\end{abstract}

\section{Introduction}

Social scientists often find themselves in situations requiring data annotation based on human judgment and specific expertise \citep{wilkerson2017large, kennedy2018gab, demszky-etal-2020-goemotions, drapal2023using}. For example, a political scientist studying hate speech on social media may need to develop a codebook that clearly defines what constitutes hate speech in the social media scenario with illustrative positive and negative examples. After developing such a codebook, the researcher may need to recruit annotators possessing sufficient domain expertise to appropriately apply the established guidelines to actual social media content.

However, both tasks---codebook development and data annotation---involve significant human effort.

Firstly, it is challenging even for domain experts to develop reliable codebook \citep{halterman2024codebook}. To start with, ambiguity and subjectivity are inherent obstacles, as interpreting complex human behaviors and communications often yields multiple valid perspectives, leading to edge cases that need specific rules to handle \citep{fornaciari-etal-2021-beyond, fuchs2021value, fleisig-etal-2023-majority, fan2025mediummessagedeconfoundingtext}. %
Even with edge case identified, the expert may struggle to explicitly articulate the subtleties and intuitions that guide their knowledge and expertise, a phenomenon commonly known as \textit{Polanyi’s paradox}\footnote{In everyday language, this phenomenon is often summarized by the phrase: ``We can know more than we can tell.''} \citep{autor2014polanyi, fugener2022cognitive}. As a result, although domain experts may possess rich implicit understandings of certain social phenomena, they face significant challenges when attempting to capture and codify this tacit knowledge within verbal, structured annotation frameworks.

Secondly, large-scale human annotation is often not feasible in many use cases \cite[e.g.,][]{xie2024automating}. Employing qualified annotators is costly, especially when the task requires domain-specific expertise. Furthermore, domain-specific data frequently involve nuanced and complex contexts \cite[e.g.,][]{ziems2024can, fan2025lexam, zhao2025scale}, which demand greater cognitive effort and longer annotation times to ensure high quality. As a result, large-scale human annotation is often prohibitively expensive, both in terms of cost and time. To address this challenge, recent research explores leveraging LLMs for automatic annotation \citep{gilardi2023chatgpt, Pangakis2023AutomatedAW, ding-etal-2023-gpt,  gpt4_vs_crowdworker, he-etal-2024-annollm, dunivin2024scalable, Trnberg2024BestPF}. These approaches typically assume the availability of well-developed codebooks for LLM prompting \citep{halterman2024codebook, xiao2023supporting}. However, how codebook development and the annotation process interact---and, crucially, how expert knowledge shapes this interaction---remains underexplored. In practice, domain experts often go back and forth between the codebook and the annotation results, updating the codebook based on insights gathered during the annotation process \cite[e.g.,][]{kirsten2025assistance}. This iterative approach is essential for discovering edge cases that the previous codebook may overlook, allowing experts to address these cases by updating the codebook accordingly.

\begin{figure*}[ht]
    \centering
    \includegraphics[width=\textwidth]{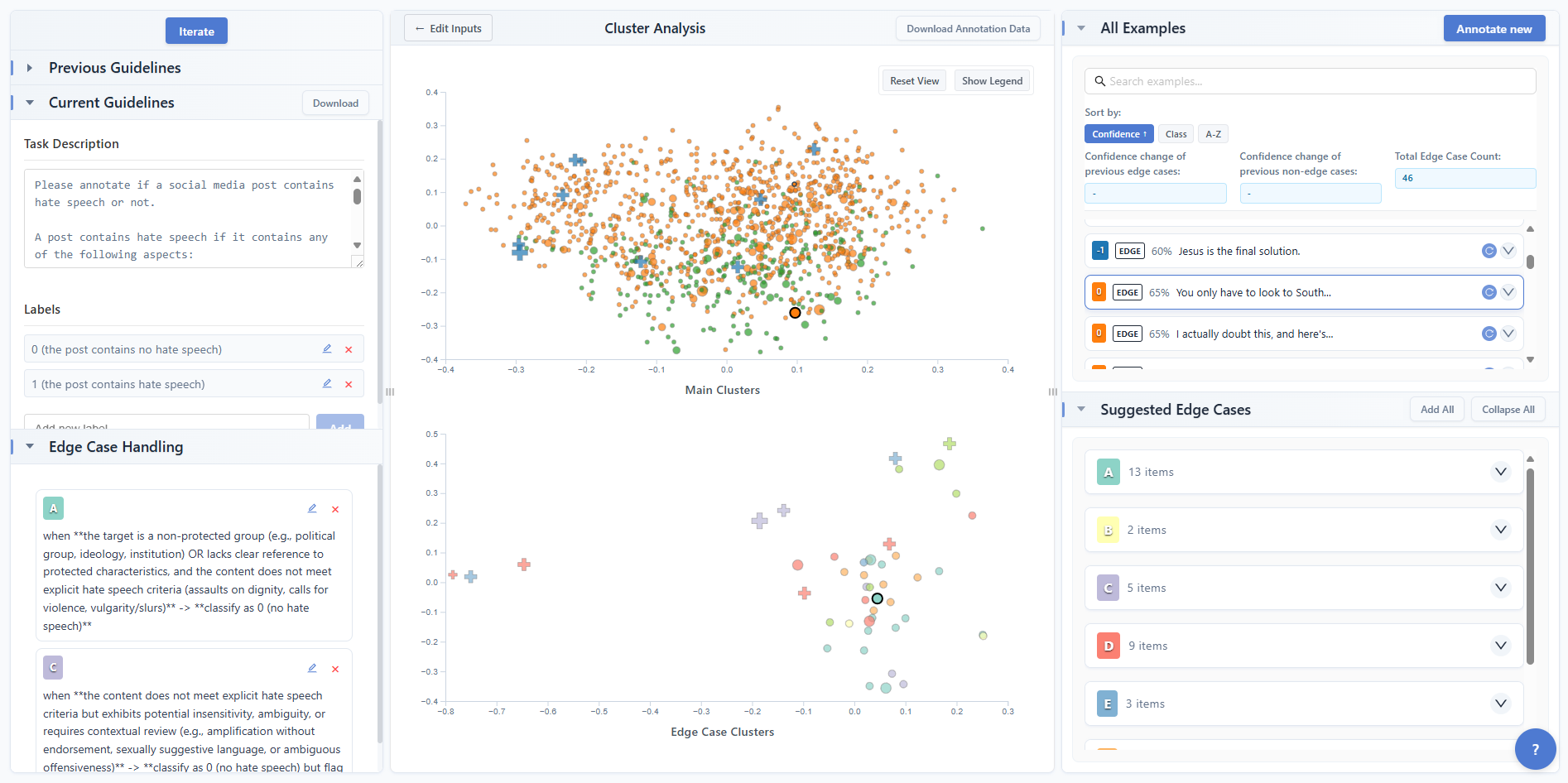}
    \vspace{-1em}
    \caption{User Interface -- Analysis Dashboard}
    \label{fig:ui_p2}
    \vspace{-1em}
\end{figure*}
To address these gaps, we introduce \systemname\space, a mixed-initiative text analysis tool designed to support human domain experts in discovering and managing edge cases, as shown in Figure \ref{fig:tf}. \systemname\space takes annotation guidelines (codebooks) and a target annotation corpus from the user as input. It then identifies edge cases (data points poorly defined by the provided codebook), strategically clusters them, and suggests aggregated edge categories along with representative examples. The user evaluates these suggested edge cases, determines their validity, and updates the codebook accordingly with clear handling rules. The user can then proceed to a new iteration of annotation using the revised codebook.

This expert-in-the-loop approach allows humans and AI to complement each other, leveraging their respective strengths through collaboration. \citet{fugener2022cognitive} and \citet{evans2011metaknowledge} show that humans struggle to identify edge cases due to a lack of \textit{metaknowledge}---the ability to assess their own capabilities and the boundaries of their own knowledge. In contrast, AIs can help to discover edge cases, while their ability to handle them is limited \citep{ni2025largelanguagemodelscapture}. It is therefore advisable to adopt an expert-in-the-loop approach, leveraging AI to discover edge cases and enrich the codebook under human supervision. %
In summary, our contributions are:
\begin{enumerate}[left=0mm,itemsep=0pt,topsep=1pt]
\item We develop \systemname\space for domain experts that iteratively updates the codebook under human supervision. Consists of an LLM-based induction algorithm suggesting representative edge cases and a user-friendly interface enabling domain experts to more effectively handle edge cases.
\item We conduct user studies with domain experts from diverse backgrounds, shedding light on the effectiveness of \systemname\space and directions for future work.
\end{enumerate}

\section{Frontend and User Work Flow} 

In this section, we provide a detailed introduction to the user workflow (illustrated in \cref{fig:tf}), including a preparation stage \cref{sec:preparation} and a dashboard analysis stage \cref{sec:dashboard}.

\subsection{Preparation Stage} \label{sec:preparation}
\myparagraph{Onboarding.} To ensure a smooth onboarding experience, the first launch of \systemname{} triggers an intro.js tour that guides users through the input and dashboard pages. Furthermore, users can also click ``Load Demo Data'' to explore a sample usecase in annotating hate speech from social media.

\myparagraph{Input Page.} After familiarized with \systemname{}, the user can start from the input page where they need to provide a initial draft of the codebook, including (1) a task definition (e.g., a post contains hate speech if it contains assaults on human dignity, calls for violence, or vulgarity.); (2) classification labels (e.g., 1 for hate speech and 0 for no hate speech); and (3) a task ID which is useful for saving all annotation outcomes, edge cases, and codebooks to the backend. Besides the codebook, the user also need to provide a csv file containing 500 to 1000 target texts to be annotated. We suggest this number of texts to ensure the representativeness of edge cases with reasonable budget. With the input prepared, the user can click ``send'' to pass the inputs to the LLM analysts. A screen shot of the input page is illustrated in Figure \ref{appendix:input_page}.

\subsection{Dashboard Analysis Stage} \label{sec:dashboard}

\myparagraph{Current Guidelines.} At the upper left-hand side of the dashboard, the user-provided codebook is displayed, allowing users to optimize their annotation task descriptions and manage labels. %

\myparagraph{Exploring Annotation Results.} At the upper middle of the dashboard, we provide a scatter plot showing all annotated text samples. Each point represents an example, clustered according to embedding similarity. Different colors indicate different annotation labels, and point size denotes annotation uncertainty---how likely the samples belong certain edge cases. Clicking on points in the scatter plot, the annotation details of the corresponding items will pop out on the upper right ``All Examples'' list, including LLM analysis, annotation confidence, and edge case suggestions.  %

\myparagraph{Analyzing Edge Cases.} Either clicking large points (uncertain annotations) in the upper scatter plot or ``Edge'' items in the upper right list will connect the user to the lower middle scatter plot and lower right list. The lower middle plot presents the clusters of potential edge cases that may require attention. These samples are automatically identified by the system as challenging or requiring more precise annotation guidelines. The panel named ``Suggested Edge Cases'' on the lower right outlines high-level descriptions of each edge case cluster, and examples in each cluster.

\myparagraph{Edge Case Handling and Iterative Optimization.} Once users find any cluster in ``Suggested Edge Cases'' reasonable, they can add the corresponding edge case handling rule to the lower left panel ``Edge Case Handling''. For example, clusters A and C are added in \cref{fig:ui_p2}. Users can also edit the edge case handling rules freely. Once they are satisfied with the added rules, they can click ``Iterate'' on the top left to re-annotate the corpus with the codebook augmented with ``Edge Case Handling''. Codebook of previous iterations will be saved in the top left panel---``Previous Guidelines''. %

\section{Backend Algorithm for Edge Case Discovery}

\begin{figure*}[ht]
    \centering
    \includegraphics[width=\textwidth]{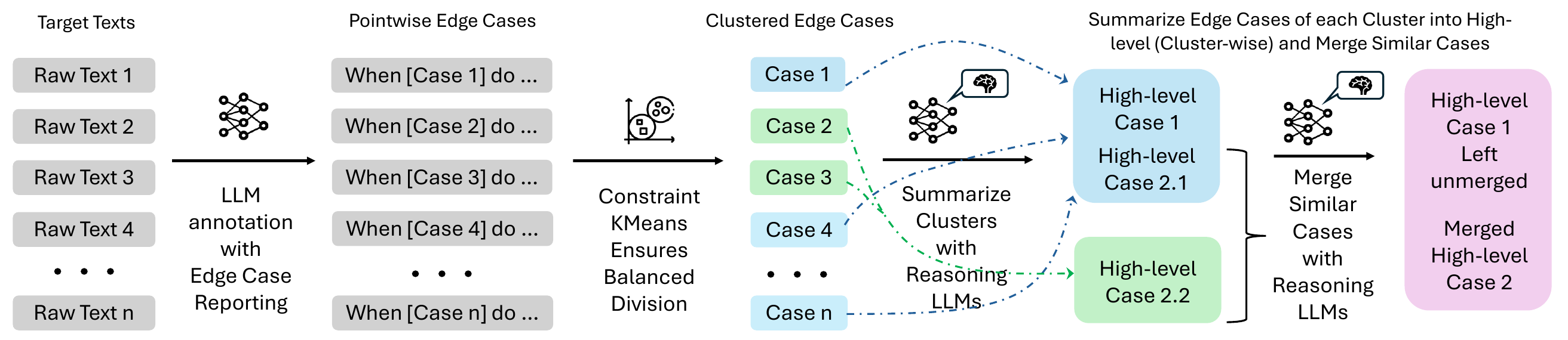}
    \vspace{-2em}
    \caption{Co-DETECT's backend algorithm for automatically discover representative edge cases. Firstly, an LLM annotator report pointwise edge cases. Secondly, a reasoning LLM aggregates item-wise edge cases into more representative high-level edge cases, with the help of clustering algorithms.}
    \vspace{-1em}
    \label{fig:backend}
\end{figure*}

\myparagraph{Problem Formulation.} As illustrated in \cref{fig:tf}, either traditional text annotation or \systemname\space requires a target corpus and task definition (i.e., initial codebook) as inputs. At this stage, the user may lack insights about the corpus, including limited edge case handlings in the codebook. \systemname\space aims at discovering edge cases that are ambiguously defined by the codebook. The edge cases proposed by \systemname\space should be:
\begin{itemize}[nosep, leftmargin=*]
    \item \textbf{Descriptive}: capture the core features of exact edge case samples and the reason why they are ambiguous.
    \item \textbf{High-Level}: while being descriptive, the edge case descriptions should not over specifically describe certain samples. Only then can they be added to the codebook and generalized to unseen data points.
\end{itemize}
To fulfill these desiderata, the edge case discovery algorithm of \systemname\space details as follows: 

\myparagraph{Step 1: Item-Level Edge Cases.} We start from using a non-reasoning LLM\footnote{We use reasoning LLMs referring to LLMs with test-time long CoT reasoning (e.g., DeepSeek-R1 \citep{deepseekai2025deepseekr1incentivizingreasoningcapability} and OpenAI O3 \citep{openai_o3_2025}); and non-reasoning LLMs referring to those answer immediately (e.g., GPT-4.1 \citep{openai_gpt4.1_2025}).} (e.g., in our case GPT-4.1) to quickly annotate all data points. We prompt the LLM to (1) annotate, (2) provide a confidence score reflecting the annotation correctness (following \citet{tian2023justaskcalibrationstrategies}), and (3) explain why the case is an edge case if the confidence is low. The explanations are in a form of edge case handling rules like ``when \texttt{[Case Description]}, do \texttt{[Action]}''. \texttt{[Case Description]} describes why the sample is ambiguously defined, so it might be too specific and low-level. \texttt{[Action]} is an LLM suggested handling for the edge case.

\myparagraph{Step 2: Cluster-Level Edge Cases.} To avoid over specific \texttt{[Case Description]} that fails to generalize to other samples, we need to aggregate item-level edge cases with similar ambiguity and describe them in a higher level. This is a challenging task requiring (1) covering all item-level edge cases; and (2) strategically finding logical similarities between reasons for ambiguity. Therefore, we employ a SOTA reasoning LLM---DeepSeek-R1 \citep{deepseekai2025deepseekr1incentivizingreasoningcapability} to cluster \texttt{[Case Description]} and generate high-level edge cases and handling rules. Specifically, we extract the item-level \texttt{[Case Description]}, embed them with semantic embedding models\footnote{In our project, we use OpenAI text-embedding-3-large \citep{openai_text_embedding_3_large_2024} for convenience.}, and cluster them with constrained KMeans \citep{Levy-Kramer_k-means-constrained_2018}. Each cluster of \texttt{[Case Description]} and corresponding \texttt{[Action]} are fed to DeepSeek-R1 to generate Cluster-wise Edge Cases. Constrained KMeans ensures that all clusters have 10 to 20 samples, so that the input (i.e., each cluster) to DeepSeek-R1 will not be too large or small, as we empirically find that large clusters ($>$20) may increase the reasoning burden and lead to hallucination, while small clusters ($<$5) may generate over-specific edge cases. Cluster-level edge cases are also companied with \texttt{[Action]} to handling them.

\myparagraph{Step 3: Merge Cluster-Level Edge Cases.} Since each cluster may have overlapped edge cases, we finally call DeepSeek-R1 again to merge cluster-level edge cases and their handling rules. This also ensures that similar edge cases are not handled with different rules.

\section{User Study}

To evaluate \systemname's effectiveness and collect feedback for further improvement, we conduct a systematic user study with domain experts. Prior to the study, participants were asked to prepare a text annotation task and an accompanying corpus from their own research domains (i.e., areas where they possess domain expertise). At the beginning of the user study, participants first complete a pre-interaction survey gathering their background information. Then, they interact with the system for approximately 45 minutes, including the LLM response time. Finally, they complete a post-interaction survey, collecting comprehensive user feedback. We recruit 10 users in total. 5 of them are not involved in the design of \systemname{}. The remaining five are co-authors of the paper but were not familiarized with the workflow before the user study.

\subsection{Pre-Interaction Survey Takeaways}
We summarize the key findings from the pre-interaction survey below. For the full survey form, please refer to~\Cref{appendix:pre}.

\myparagraph{Diverse Experience and Background of the Participants.} Our participants have a broad academic background in social science, computational linguistics, and interdisciplinary training. They also exhibit diverse experience in both social science qualitative coding and LLM-assisted annotation, from no experience to expert level. 

\myparagraph{Heavy Reliance on Manual Effort for Edge Case Discovery.} Concerning common workflows for identifying edge cases, 80\% of participants manually review subsets of data to detect potential edge cases. Some also report employing other human (e.g., crowdworkers) or AI annotators to do pilot annotation and flag potential edge cases.

\myparagraph{Moderate Prior Knowledge of Edge Cases.} 70\% of participants report knowing certain edge cases in their intended datasets. Therefore, it would be valuable to check if \systemname{} can mine already-known edge cases or discover new edge cases.

\subsection{Post-Interaction Survey Takeaways}
Below, we highlight the main insights from the post-interaction survey, which center on four key aspects of user experience with \systemname{}: ease of use, interpretability of visualizations, validity of edge cases, and overall satisfaction and feedback. For the full survey form, please refer to~\Cref{appendix:post}.

\myparagraph{The Majority Finds \systemname{} Workflow Easy to Follow.} The survey results reveal generally positive feedback on interface ease of use and task clarity, with most participants (80\%) finding navigation intuitive and interaction straightforward. Some requested additional visualizations (e.g., density distribution of the confidence scores) or export features for enhanced usability.

\myparagraph{\systemname{} Can Identify Relevant Edge Cases.} 60\% of participants approved \systemname{}'s ability to clearly identify relevant edge cases. For example, one participant reported that the edge case handling rules suggested are ``clear, realistic, and concise'', and directly inform their acceptance or rejection decisions, pointing out ways to improve the precision and coverage of these suggestions. 90\% of participants report that \systemname{} may help discover new edge cases beyond their prior knowledge of the dataset.

\myparagraph{Useful Iterative Workflow and Overall Satisfaction.} 80\% of participants find the iterative feature of \systemname{} useful for refining their annotation guidelines. All participants were satisfied with the Co-DETECT system's support in generating annotation guidelines and identifying new edge cases.

\myparagraph{Constructive Critiques.} Besides the generally positive feedback on user experience of \systemname{}, the post-interaction survey also gives us valuable critiques, highlighting areas for future improvement of \systemname{}. 40\% of participants express concern that \systemname{} may overlook potential edge cases although the identified edge cases seem reasonable. For instance, one participant also indicated that there was room to enhance the coherence and descriptive clarity of the edge cases.

\subsection{Quantitative Human Evaluation on Edge Case Validity}

We further conducted a quantitative human evaluation with three participants\footnote{Due to the original user study is already time-intensive, participation in the quantitative evaluation was optional.}. Each participant was asked to randomly select 1 to 2 samples from each edge case cluster and manually assess how many were accurately captured by the \systemname{}-suggested edge case descriptions. Among 41 randomly selected samples, 33 (\textbf{80.5\%}) were reported as well-described by the suggested edge case descriptions. The edge case descriptions are also found to be sufficiently high-level to cover more than one samples. We further find that it often takes less than 5 seconds for an expert to identify if a sample is covered by an edge case description or not, indicating that \systemname{} may not impose a heavy cognitive load on users when supervising suggested edge case clusters.

\section{Can Improved Codebook Benefit Automatic Annotation?}

\begin{table}[t]
\centering
\begin{tabular}{lcc}
\toprule
\textbf{Dataset} & \textbf{1st Iter.} & \textbf{2nd Iter.} \\ 
\midrule
GabHateCorpus & 0.2144 & \textbf{0.2523} \\
GoEmotions-Positive & 0.0300 & \textbf{0.3297} \\
GoEmotions-Negative & 0.2823 & \textbf{0.3046} \\
\bottomrule
\end{tabular}
\caption{Classification F1 Scores using the original codebook (from the 1\textsuperscript{st} iteration) and the improved codebook after one \systemname{} iteration (from the 2\textsuperscript{nd} iteration).}
\label{tab:f1_scores}
\end{table}

The main goal of \systemname{} is to help experts improve codebooks, but does a better codebook actually enhance automatic annotation? To investigate this, we provide GPT-4.1 with codebooks before and after \systemname{} enhancement and compare its classification F1, varying only the codebook. We strategically pick a hate speech detection dataset---GabHateCorpus \citep{kennedy2021gabhatecorpus} and an emotion classification---GoEmotions (\citealp{demszky-etal-2020-goemotions}; Positive / Negative Emotion Detection) for this evaluation, because these tasks are highly subjective \citep{10.1162/tacl_a_00449,ni2025largelanguagemodelscapture} and thus challenging for codebook drafting. They are therefore challenging for advanced LLMs like GPT-4.1 that are smart enough to understand the nuanced perturbations within the codebook. It is also challenging for human experts to manually improve codebook as it is hard for individuals to capture various subjectivity.

The results are exhibited in \cref{tab:f1_scores}, where we observe an increase in F1 scores across different datasets. Notably, \systemname{} only augments codebook by appending edge case handling rules. The initial codebook for GoEmotion-Positive has very low F1 score due to an extremely low recall---the model rarely predicts positive emotions that are not explicitly stated by the raw codebook. Thereby, we showcase that \systemname{} can improve classification outcomes with improved codebook, even for subjective tasks that are both challenging for LLMs and individual experts.

\section{Related Work}
\myparagraph{Annotation with LLM Assistance.} Both NLP \citep{kim-etal-2024-meganno,ni-etal-2024-afacta,ni-etal-2025-diras} and HCI \citep{gpt4_vs_crowdworker,törnberg2023chatgpt4outperformsexpertscrowd} community have widely explored human-AI collaborations for text classification. \citet{tian-etal-2023-just} find that the verbalized confidence of LLMs indicates classification quality, and annotations where LLM reports high annotation confidence may outperform human annotator \citep{ni-etal-2024-afacta,ni-etal-2025-diras,törnberg2023chatgpt4outperformsexpertscrowd}. We follow this stream of work to calibrate the quality of LLM annotation using verbalized confidence scores. In Human-AI interaction, \citet{human_ai_collaborative_annotation} and \citet{kim-etal-2024-meganno} develop mixed-initiative tools to enhance automatic annotation with minimal human supervision. However, these methods focus on annotation accuracy and assume a predefined codebook. In contrast, our work targets efficiency in codebook development and edge-case discovery, critical steps especially in the initial stages of text classification \citep{Trnberg2024BestPF}.

\myparagraph{Goal-Driven Clustering in NLP.} One critical step of our edge case discovery algorithm is to cluster low-level specific edge cases into high-level representative edge cases. Such goal-driven clustering \citep{wang2023goaldrivenexplainableclusteringlanguage} is essentially relevant to many NLP sub-fields, such as topic modeling \citep{pham-etal-2024-topicgpt}, inductive reasoning \citep{Lam_2024}, corpus comparison \citep{D5_zhong_2023}, information retrieval \citep{ni-etal-2025-diras} etc. In such tasks, LLM plays an important role in understanding users' goal and steering / interpreting the clustering accordingly \citep{zhang-etal-2023-clusterllm,viswanathan-etal-2024-large,movva2025sparseautoencodershypothesisgeneration}. Our work contributes to adapting goal-driven clustering to edge case discovery, leveraging analytical skills of reasoning models \citep{deepseekai2025deepseekr1incentivizingreasoningcapability}.

\section{Conclusion}

We developed \systemname{} to systematically identifies descriptive and generalizable edge cases and collaboratively improve codebook with human expert. To achieve this, \systemname{} induces representative edge cases leveraging multi-step clustering and reasoning LLMs. Then the user can supervise the quality of suggested edge cases and decide whether to include them into the codebook or not. Comprehensive user study, and other qualitative and quanititative evaluations prove the effectiveness of \systemname{}. %

\section*{Ethics Statement}

This research involved voluntary participation in user studies, during which participants provided professional background information and evaluated interface functionality for annotation and edge-case identification tasks. Participants were clearly informed about the study objectives, tasks, and their right to withdraw at any time. Collected data were securely stored, anonymized, and analyzed collectively to ensure confidentiality and privacy. The study posed minimal risks to participants, aligned with standard professional activities, and adhered closely to ethical guidelines for human-centered research.

\section*{Broader Impact Statement}

Our system pipeline emphasizes an interactive and iterative approach designed to enhance annotation accuracy and generalization through the systematic management of challenging edge cases. This approach is based on the assumption that improved confidence metrics in a model correlate with enhanced annotation performance. Nevertheless, analogous to the \textit{Clever Hans} phenomenon---where an intelligent system identifies unintended cues instead of genuinely learning underlying knowledge---it is crucial to critically assess the robustness of this pipeline against potential biases and unintended shortcuts that may result from repeated feedback loops and rule-induction processes.

One potential concern involves deriving edge-case rules primarily from model confidence metrics and automatically suggested edge-case instances. If the model's selection and clustering of these edge cases rely predominantly upon internally generated confidence measures, there is a risk that inductively derived rules may reinforce model-specific biases rather than reflect genuinely generalizable conceptual regularities. For example, the model might inadvertently identify clusters based on spurious correlations between input texts and target labels instead of addressing genuine annotation challenges.

Therefore, such risks necessitate increased caution. Systems optimized exclusively within internal frameworks may achieve superficially impressive performance improvements without truly acquiring underlying domain expertise. Consequently, we emphasize the critical importance of human domain expertise. Users should ensure that only rules verified by domain experts are included when updating the codebook.

\section*{Acknowledgment}

We would like to express our gratitude to \textit{Cong Jiang, Felix Riechmann, Yang Tian, Tingyu Yu,} and \textit{Darya Zare} for their generous support throughout this study.

\bibliography{anthology, custom}

\appendix

\begin{figure*}[ht]
    \centering
    \includegraphics[width=\textwidth]{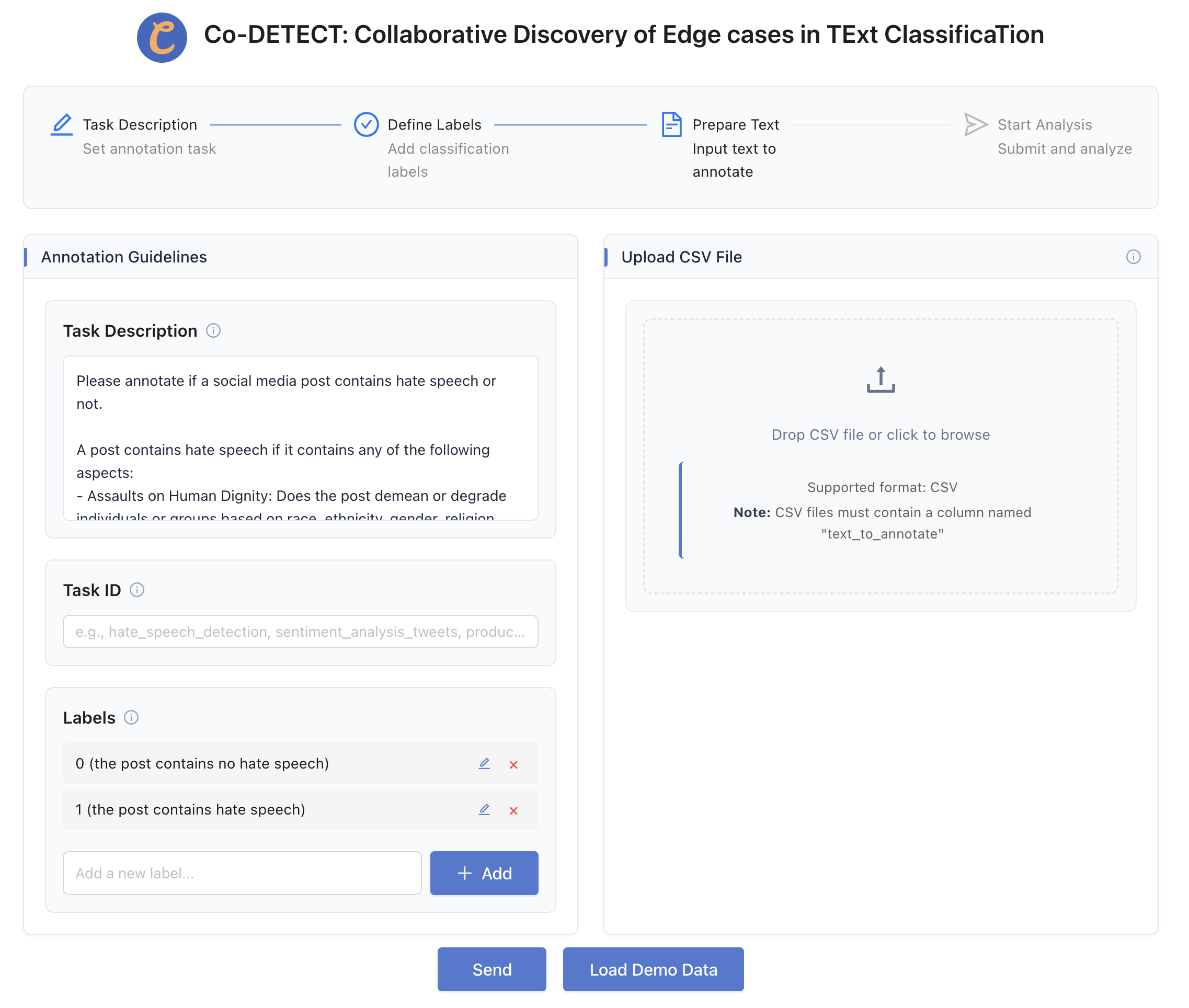}
    \caption{User Interface -- Homepage.}
    \label{fig:ui_p1}
\end{figure*}
\label{appendix:input_page}

\section{Pre-Interaction Survey Questions}
\label{appendix:pre}

In the pre-interaction survey, we included following questions:

\myparagraph{(1) General User Background}

\begin{itemize}
    \item Please briefly describe your professional/academic background and current research areas.
    \item What is your previous experience with data annotation and qualitative coding? \textit{(None, Beginner, Intermediate, Advanced, Expert)}
    \item Have you used annotation support or guideline-generation tools previously? \textit{(Yes, No)}
\end{itemize}

\myparagraph{(2) Expectation}

\begin{itemize}
    \item What is your normal workflow of identifying edge cases in text annotation?
    \item In the dataset that you plan to analyze with AutoDETECT, did you already know any edge cases? \textit{(Yes, No)}
\end{itemize}

\section{Post-Interaction Survey Questions}
\label{appendix:post}

In the post-interaction survey, we included following questions:
\\
\myparagraph{(1) Task Completeness, Clarity, and Ease of Use}

\begin{itemize}

\item Were you clearly able to understand the steps for configuring an annotation task using the interface? \textit{(Five-level Likert item, strongly agree to strongly disagree)}

\item Did you encounter any difficulty navigating through different interface components (homepage to analysis dashboard)? \textit{(Yes, No)}

\item If yes, please briefly explain your difficulty.

\end{itemize}

\myparagraph{(2) Annotation Results Visualization and Interpretation}

\begin{itemize}

\item Were you able to clearly interpret and interact with the annotation results displayed in the scatter plots ("All Examples" and "Suggested Edge Cases")? \textit{(Five-level Likert item, strongly agree to strongly disagree)}

\item Did interactive features (e.g., clicking points to highlight examples across lists and plots) support your understanding of annotation results effectively? \textit{(Five-level Likert item, supports fully to doesn't support at all)}

\item Would you prefer alternative ways of visualizing or interacting with annotation results visually? \textit{(Yes, No)}

\item If yes, please describe briefly.

\end{itemize}

\myparagraph{(3) Edge Case Identification and Handling}

\begin{itemize}

\item Was the component provided by the system clearly identifying relevant and helpful edge cases (cases that require additional annotation guidance) in your corpus? \textit{(Five-level Likert item, strongly agree to strongly disagree)}

\item Do the edge cases make sense? \textit{(Five-level Likert item, makes total sense to makes absolutely no sense)}

\item Are the proposed rules easy to follow? \textit{(Five-level Likert item, very easy to very difficult)}

\item Please describe briefly your reasoning for accepting or rejecting suggested edge-case rules. What information or criteria were the most important for your decisions?

\end{itemize}

\myparagraph{(4) Iterative Optimization Support}

\begin{itemize}

\item Did you find the iterative approach ("Iterate" button functionality) helpful for progressively refining your annotation guidelines and labels? \textit{(Five-level Likert item, very helpful to not helpful at all)}

\item How many iterations (approximately) did you perform? Did subsequent iterations help significantly in clarifying your annotation guidelines? Please briefly explain.

\end{itemize}

\myparagraph{(5) General User Experience and Satisfaction}

\begin{itemize}

\item Did Co-DETECT help you to find some new edge cases that you didn’t notice before? \textit{(Yes, No, Maybe)}

\item How satisfied are you overall with the functionality that this tool offers you in developing codebooks and annotation guidelines? \textit{(Five-level Likert item, very satisfied to not satisfied at all)}

\item Do you still have concern that e.g. there are missing edge cases not identified by the system? \textit{(Yes, No, Maybe)}

\item What features, if any, do you think are missing or need improvement in this tool?

\end{itemize}

\myparagraph{(6) Open-ended Feedback and Improvement Suggestions}

\begin{itemize}

\item What did you like the most about the user interface and its functionality?

\item What improvements or additions would you propose to enhance the usability or functionality of the current interface?

\item (Optional) Any additional comments or suggestions about the tool or your experience using it?

\end{itemize}

\end{document}